\documentclass[letterpaper]{article} 
\usepackage{aaai23}  
\usepackage{times}  
\usepackage{amsfonts}
\usepackage{helvet}  
\usepackage{courier}  
\usepackage[hyphens]{url}  
\usepackage{graphicx} 
\usepackage{natbib}  
\usepackage{caption} 
\usepackage{algorithm}
\usepackage{algorithmic}
\usepackage{multirow}
\usepackage{booktabs}
\usepackage{bbding}
\usepackage{amsmath}
\usepackage{caption}
\usepackage{subcaption}

\usepackage{enumitem}
\usepackage{newfloat}
\usepackage{listings}
\DeclareCaptionStyle{ruled}{labelfont=normalfont,labelsep=colon,strut=off} 
\floatstyle{ruled}
\newfloat{listing}{tb}{lst}{}
\floatname{listing}{Listing}
\title{Improving Interpretability of Deep Sequential Knowledge Tracing Models with Question-centric Cognitive Representations}
\author {
    Jiahao Chen\textsuperscript{\rm 1},
    Zitao Liu\textsuperscript{\rm 2}\thanks{The corresponding author: Zitao Liu},
    Shuyan Huang\textsuperscript{\rm 1},
    Qiongqiong Liu\textsuperscript{\rm 1},
    Weiqi Luo\textsuperscript{\rm 2}
}
\affiliations {
    \textsuperscript{\rm 1} TAL Education Group, Beijing, China \\   \textsuperscript{\rm 2} Guangdong Institute of Smart Education, Jinan University, Guangzhou, China\\
    chenjiahao@tal.com, liuzitao@jnu.edu.cn, huangshuyan@tal.com, liuqiongqiong1@tal.com, lwq@jnu.edu.cn
}

\usepackage{bibentry}

\begin{document}

\maketitle

\begin{abstract}

Knowledge tracing (KT) is a crucial technique to predict students' future performance by observing their historical learning processes. Due to the powerful representation ability of deep neural networks, remarkable progress has been made by using deep learning techniques to solve the KT problem. The majority of existing approaches rely on the \emph{homogeneous question} assumption that questions have equivalent contributions if they share the same set of knowledge components. Unfortunately, this assumption is inaccurate in real-world educational scenarios. Furthermore, it is very challenging to interpret the prediction results from the existing deep learning based KT models. Therefore, in this paper, we present QIKT, a question-centric interpretable KT model to address the above challenges. The proposed QIKT approach explicitly models students' knowledge state variations at a fine-grained level with question-sensitive cognitive representations that are jointly learned from a question-centric knowledge acquisition module and a question-centric problem solving module. Meanwhile, the QIKT utilizes an item response theory based prediction layer to generate interpretable prediction results. The proposed QIKT model is evaluated on three public real-world educational datasets. The results demonstrate that our approach is superior on the KT prediction task, and it outperforms a wide range of deep learning based KT models in terms of prediction accuracy with better model interpretability. To encourage reproducible results, we have provided all the datasets and code at \url{https://pykt.org/}.




\end{abstract}

\section{Introduction}
\label{sec:intro}

\begin{figure}[!tbph]
\centering
\includegraphics[width=\columnwidth]{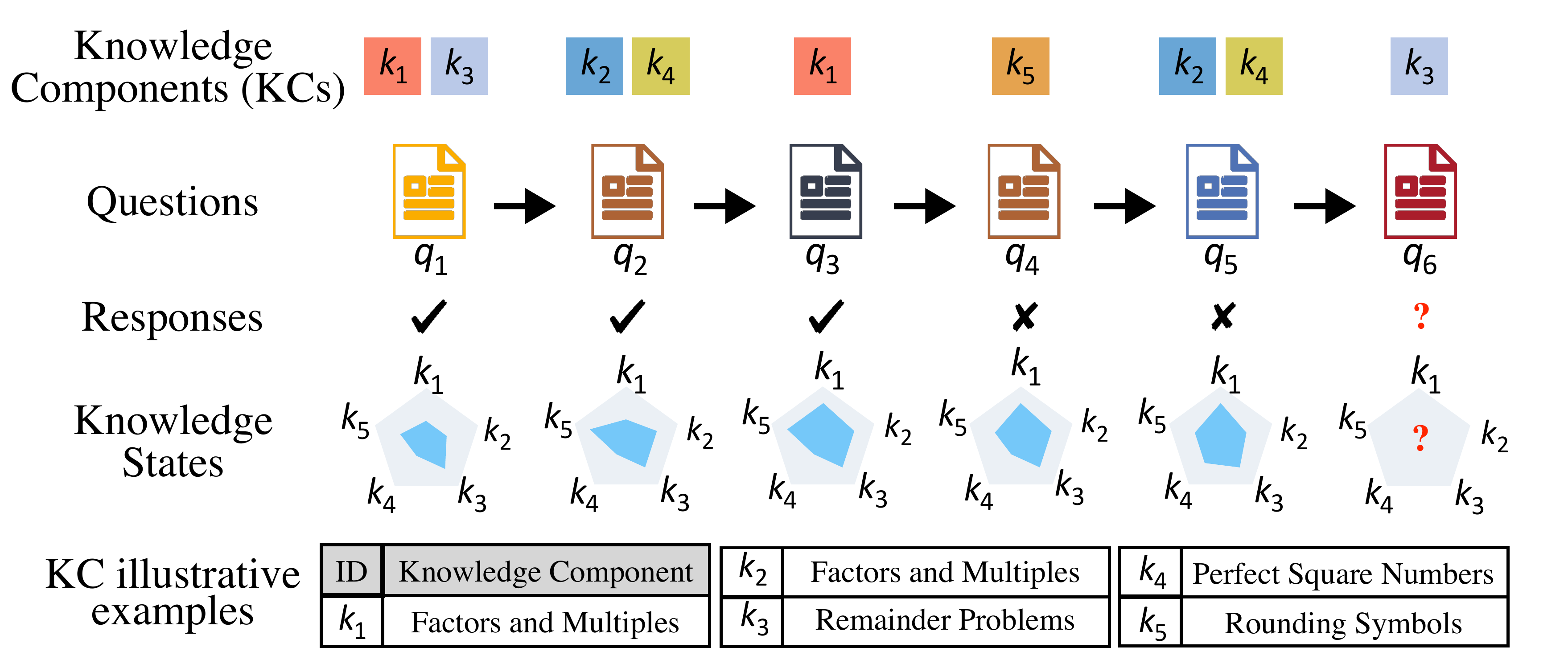}  \vspace{-0.4cm}
\caption{A graphical illustration of the KT problem.}
\label{fig:kt_illustration}
\vspace{-0.6cm}
\end{figure}

Knowledge tracing (KT) is the task of using students' historical learning interaction data (e.g., responses to a series of questions) to model their knowledge mastery over time so as to make predictions on their future performance (e.g., predicting correctly on next question) \cite{corbett1994knowledge}. Figure \ref{fig:kt_illustration} gives an illustrative example of the KT task. Such predictive capabilities can potentially help students learn better and faster when paired with high-quality learning materials and instructions and the KT models have been widely used to support intelligent tutoring systems and MOOC platforms \cite{kaser2017dynamic,cen2006learning,lavoue2018adaptive,liu2021mathematical}.

Recently, remarkable progress has been made by applying deep learning techniques to solve the KT problem \cite{piech2015deep,abdelrahman2019knowledge,ghosh2020context,nakagawa2019graph,pandey2019self,pandey2020rkt,shen2021learning,shen2020convolutional,yang2020gikt,zhang2017dynamic,zhang2021multi,wang2019deep,liu2023enhancing,liu2023simplekt}. One of the representative approaches among them is the deep sequential KT modeling, which utilizes auto-regressive architectures, such as LSTM and GRU, to represent student's knowledge states (e.g., the mastery level of the concepts) as the hidden states of recurrent units \cite{piech2015deep,chen2018prerequisite,guo2021enhancing,lee2019knowledge,liu2019ekt}. Due to the ability to learn sequential dependencies from student interaction data, deep sequential KT models draw attention from researchers from different communities and achieve great success in improving the KT prediction accuracy \cite{minn2018deep,nagatani2019augmenting,su2018exercise,yeung2018addressing}.

In spite of the promising results demonstrated by previous methods, some important limitations still exist when applying deep sequential KT models on real-world educational data. First, most existing approaches rely on the \emph{homogeneous assumption} that questions nested under a particular set of knowledge components (KCs) are equivalent \cite{zhang2017dynamic,nagatani2019augmenting,nakagawa2019graph,lee2019knowledge}. The homogeneous assumption is inaccurate in two perspectives: (1) it assumes that students have the same knowledge increment after they give the same responses to homogeneous questions\footnote{In this paper, we refer to questions that have the same set of KCs as ``\emph{homogeneous questions}''.} during the knowledge acquisition learning processes; and (2) it assumes that students will give the same responses to different questions as long as these questions are homogeneous during the problem solving process. Such unrealistic assumption limits the KT performance of the previous works. While in some cases the problem may be alleviated by implicitly modeling the question difficulty or question discrimination \cite{zhang2021multi,liu2021improving,ghosh2020context}, they suffer from the lack of ground truth labels or the exclusions of cognitive modeling (e.g., only used in pre-trained tasks), and jointly modeling the question-centric cognitive effects on knowledge states remains a big concern. Second, although deep learning based knowledge tracing (DLKT) models have shown advanced progress in terms of prediction accuracy compared with traditional cognitive models, it is difficult to extract psychologically meaningful explanations from their million-level parameters, that would relate to cognitive theory. The lack of nontransparent decision processes of DLKT models is unsatisfied for tutors and students who need to see a convincing diagnosis before they accept results generated from DLKT models.

In this paper, we address aforementioned challenges by proposing a novel KT model called \emph{Q}uestion-centric \emph{I}nterpretable \emph{K}nowledge \emph{T}racing, i.e., \emph{QIKT}. More specifically, QIKT explicitly learns question-centric cognitive representations with a knowledge acquisition module and a problem solving module. The knowledge acquisition module aims to model the variations in students' knowledge states after receiving responses to specific questions. It estimates students' question-specific knowledge acquisition by a joint optimization including representations of students' current knowledge states, responses, questions and the corresponding KCs. The problem solving module estimates students' problem solving abilities on each specific question by projecting student knowledge states on the jointly learned representations of both questions and KCs. Furthermore, the QIKT incorporates an interpretable prediction layer to improve interpretability of prediction results. The interpretable prediction layer is built upon the Item Response Theory (IRT) in psychometrics, and integrates the parameters of an IRT model into the question-centric deep sequential KT model. This enables the QIKT model to generate explainable personalized parameters for each student at question level. We evaluate QIKT on three benchmark datasets by comparing it with 13 previous approaches under a rigorous  KT evaluation protocol \cite{liu2022pykt}. Experimental results demonstrate that QIKT achieves superior prediction performance and the psychologically meaningful interpretability simultaneously.

The main contributions are summarized as follows:

\begin{itemize}[leftmargin=*]
\item We introduce a knowledge acquisition module and a problem solving module to learn question-centric representations when students absorb knowledge after answering questions and apply knowledge to solve problems.

\item We design a simple yet effective interpretable prediction layer based on the IRT theory and manage to seamlessly combine it with existing deep sequential KT models.

\item We conduct comprehensive quantitative and qualitative experiments to validate the performance of QIKT on three public datasets with a wide range of baselines. The well-designed experiments illustrate the superiority of our approach in both prediction performance and model interpretability. To the best of our knowledge, \textbf{our QIKT model is able to achieve the best prediction performance in terms of AUC} on the publicly available reproducible KT experimental settings. 
\end{itemize}

\section{Background and Related Work}

\subsection{Deep Sequential Modeling for Knowledge Tracing}
\label{sec:bg_dkt}

Deep sequential KT models utilize an auto-regressive architecture to capture the intrinsic dependencies among students' chronologically ordered interactions \cite{chen2018prerequisite,guo2021enhancing,lee2019knowledge,liu2019ekt,minn2018deep,nagatani2019augmenting,piech2015deep,su2018exercise,yeung2018addressing}. Since the very first and successful research work of deep knowledge tracing (DKT) that applies recurrent neural networks to model students' dynamic learning behaviors by \citet{piech2015deep}, a large number of works have been done to improve DKT's performance \cite{yeung2018addressing,chen2018prerequisite,su2018exercise,nagatani2019augmenting,lee2019knowledge,liu2019ekt,guo2021enhancing}. For example, \citet{yeung2018addressing} proposed to use two regularization terms to address the reconstruction and waviness issues in the DKT model. \citet{chen2018prerequisite} incorporated prerequisite relations between pedagogical concepts to enhance DKT model. \citet{su2018exercise} presented to aggregate textual representations to monitor student knowledge states. \citet{nagatani2019augmenting} developed approaches to capture students' forgetting behaviors and  \citet{guo2021enhancing} leveraged adversarial training samples to enhance the deep sequential KT models' generalization.


Besides the deep sequential KT models, other types of neural network based approaches are applied in the KT domain as well, such as memory augmented KT models that explicitly model latent relations between KCs with an external memory \cite{abdelrahman2019knowledge,shen2021learning,zhang2017dynamic}, graph based KT models that capture interaction relations with graph neural networks \cite{nakagawa2019graph,tong2020structure,yang2020gikt}, and attention based KT models that use the attention mechanism and its variants to capture dependencies between interactions \cite{ghosh2020context,pandey2020rkt,pu2020deep,zhang2021multi}.

\subsection{Interpreting Deep Learning Based Knowledge Tracing Models}
\label{sec:related}
\begin{figure*}[!bpht]
\centering
\includegraphics[width=\textwidth]{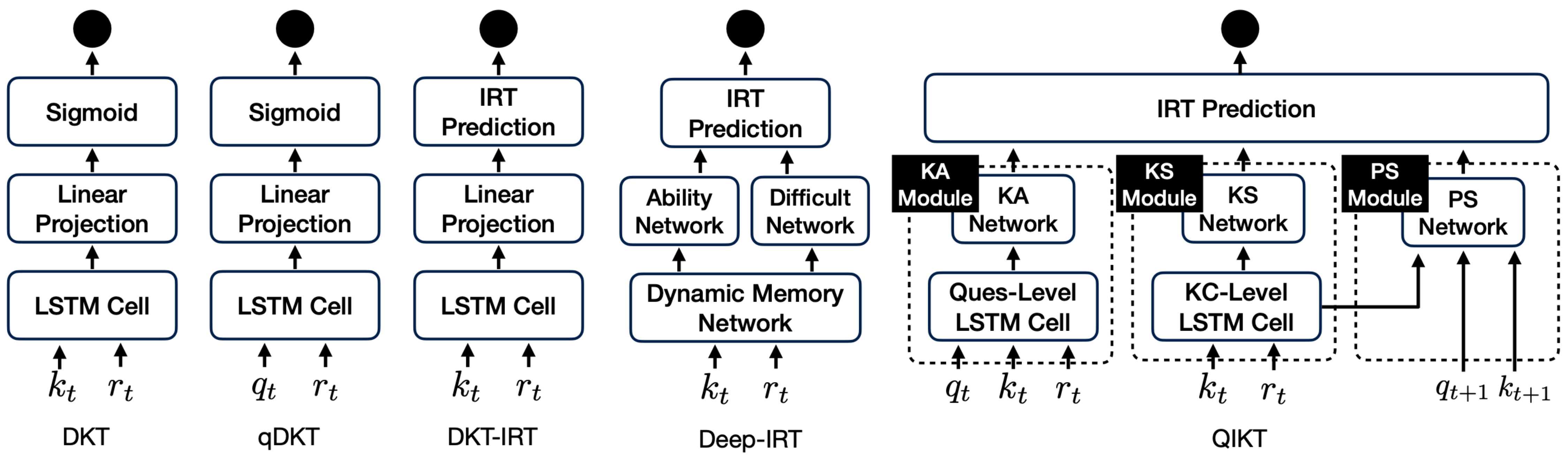}  \vspace{-0.7cm}
\caption{Graphical illustrations of our QIKT model along with some representative DLKT models including DKT \cite{piech2015deep}, qDKT \cite{sonkar2020qdkt}, DKT-IRT \cite{converse2021incorporating}, and Deep-IRT \cite{yeung2019deep}.} \vspace{-0.4cm}
\label{fig:framework}
\end{figure*}

Recently, many interpretable approaches have been incorporated into DLKT models for both student modeling and prediction tasks. These techniques can be divided into the following three categories:

\begin{itemize}[leftmargin=*]
\item \textbf{C1: Post-hoc local explanation}. Post-hoc local explanation techniques are used in the KT task aiming to examine each individual prediction result and figure out why the DLKT models make the decisions they make \cite{lu2020towards,lu2022interpreting}. For example, \citet{lu2022interpreting} applied a layer-wise relevance propagation method to interpret a deep sequential KT model by back propagating relevance scores from the model's output layer to its input layer. 

\item \textbf{C2: Global interpretability with explainable structures}. Embed an interpretable cognitive module into existing DLKT architectures to better understand the knowledge state modeling process \cite{wang2020neural,zhao2020interpretable,pu2022eakt}. For example, \citet{wang2020neural} designed an intermediate interaction layer based on multidimensional IRT and explicitly modeled both student factors and exercise factors. \citet{pu2022eakt} proposed to utilize an automatic temporal cognitive method to better capture the changes in students' knowledge states.

\item \textbf{C3: Global interpretability with explainable parameters}. Directly use cognitively interpretable models to estimate the probability that a student will answer a question correctly. Explainable parameters in these models are obtained from outputs of the DLKT models \cite{converse2021incorporating,yeung2019deep}. For example, \citet{converse2021incorporating} linearly transformed the hidden states of the DKT model and then applied a hard thresholding operator to cast the parameters into the IRT-like form. \citet{yeung2019deep} proposed to explicitly learn levels of student abilities and KC difficulties with a dynamic key-value memory network for KT task and feed the learned results to an IRT layer for final prediction. 

\end{itemize}

Our QIKT approach belongs to the C3 category due to the fact that we utilize the IRT function as the final layer for interpretable prediction. Different from existing approaches \cite{yeung2019deep,converse2021incorporating} that only optimize the model performance based on interpretable predicted outcomes, our QIKT approach directly incorporates the explainable parameter learning into the final model optimization objective, which improves the model interpretability and preserves the prediction performance as well.  Compared with the methods developed based on memory networks such as Deep-IRT \cite{yeung2019deep}, our approach is based on the deep sequential architectures, which is more applicable and has better prediction accuracy \cite{liu2022pykt}.


\section{Problem Statement}
\label{sec:ps}

Our objective is given an arbitrary question $q_*$ to predict the probability of whether a student will answer $q_*$ correctly based on the student's historical interaction data. More specifically, for each student $\mathbf{S}$, we assume that we have observed a chronologically ordered collection of $t$ past interactions i.e., $\mathbf{S} = \{\mathbf{s}_j\}_{j=1}^t$. Each interaction is represented as a 4-tuple $\mathbf{s}$, i.e., $\mathbf{s} = <q, \{k | k \in \mathcal{N}_q \}, r, s>$, where $q, \{k\}, r, s$ represent the specific question, the associated KC set, the binary valued student response\footnote{Response is a binary valued indicator variable where 1 represents the student correctly answered the question, and 0 otherwise.}, and student's response timestamp respectively. $\mathcal{N}_q$ is the set of KCs that are associated with question $q$. We would like to estimate the probability $\hat{r}_{*}$ of the student's future performance on arbitrary question $q_*$.

\section{Interpretable KT Modeling with Question-centric Cognitive Representations}
\label{sec:method}
In this section, we provide details about our QIKT model that is made up of five components: (1) the interaction encoder that assembles and encodes both question-level and KC-level information; (2) the question-centric knowledge acquisition (KA) module that examines students' knowledge acquisition after answering specific questions over time; (3) the question-agnostic knowledge state (KS) module that models the general knowledge state dynamics; (4) the question-centric problem solving (PS) module that estimates the capabilities of students to tackle a specific question with their current knowledge states; and (5) the interpretable prediction layer that aims to leverage the psychological theory of IRT to generate more interpretable results for both tutors and students.

\subsection{Interaction Encoder}
\label{sec:encoder}
\begin{figure*}[!bpht]
\centering
\includegraphics[width=\textwidth]{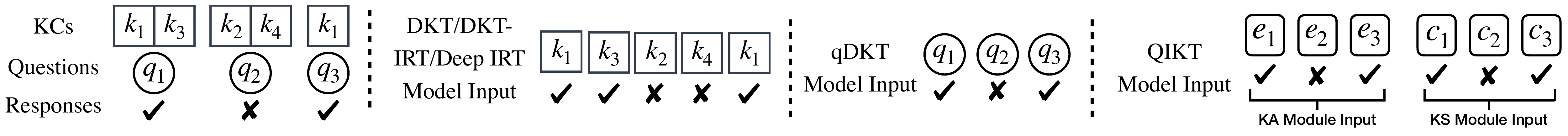}  \vspace{-0.6cm}
\caption{Illustrations of different interaction encoding approaches. $\mathbf{e}_i$s and $\mathbf{c}_is$ are defined in eq.(\ref{eq:et}) and eq.(\ref{eq:ct}) respectively.} \vspace{-0.5cm}
\label{fig:encoder_illustration}
\end{figure*}


In real-world educational scenarios, the question bank is usually much bigger than the set of KCs, for example, the number of questions is more than 1500 times larger than the number of KCs in the well cited Algebra2005 dataset\footnote{Details of Algebra2005 is described in the \emph{Datasets} section.} (see Table \ref{tab:sta}). Therefore, most existing research works like DKT \cite{piech2015deep}, DKT-IRT \cite{converse2021incorporating}, and Deep-IRT \cite{yeung2019deep} alleviate the data sparsity issue by using KCs to index questions and all questions cover the same KC are treated as a single question. Although this conversion greatly relieves the sparsity problem \cite{ghosh2020context,liu2022pykt}, it enforces the DLKT models to follow the homogeneous assumption and hence omits different learning effects brought by questions of the same concepts. Please note that this conversion lets the DLKT models be learned from the extended KC-level sequences instead of the original question-level sequences, as illustrated in Figure \ref{fig:encoder_illustration}. On the other hand, question-centric models like qDKT \cite{sonkar2020qdkt} completely ignore the relations between questions and KCs and purely uses the question sequence to track students' knowledge states.

In this work, we aim to improve the aforementioned DLKT models by capturing the intrinsic relations between questions and KCs at a more fine-grained level. More specifically, we have two different raw interaction encodings for the KA module and the KS module. For the KA module, similar to a recent work by \citet{long2021tracing}, each question level interaction $\mathbf{e}_t$ is represented as a combination of question, response and the corresponding set of KCs, i.e., 

\vspace{-0.3cm}

\begin{align}
	\label{eq:et}
	\mathbf{e}_{t}=\left\{
		\begin{array}{l}
			\mathbf{q}_t \oplus \bar{\mathbf{k}}_t \oplus \mathbf{0}, \quad r_{t}=1 \\
			\mathbf{0} \oplus \mathbf{q}_t \oplus \bar{\mathbf{k}}_t, \quad  r_{t}=0 \\
		\end{array}\right.
\end{align}

\noindent where $\mathbf{q}_t$ is the question embedding, $\mathbf{q}_t \in \mathbb{R}^{d \times 1}$ and $\bar{\mathbf{k}}_t$ is the average embeddings of all the associated KCs to the $t$th question, i.e., 

\vspace{-0.4cm}

\begin{equation*}
	\bar{\mathbf{k}}_t = \frac{1}{\left|K_{q_t}\right|} \sum_{j = 1}^m \mathbf{k}_j * \mathbb{I}(k_j \in \mathcal{N}_{q_t})
\end{equation*}

\noindent where $\mathbf{k}_j$ is the KC embedding, $\mathbf{k}_j \in \mathbb{R}^{d \times 1}$. $m$ is the total number of KCs in the question bank. $K_{q_t}$ is the size of $\mathcal{N}_{q_t}$. $\mathbb{I}(\cdot)$ is the indicator function and $\oplus$ is the concatenate operation. The response in each interaction is encoded as a $2d \times 1$ all-zero vector, $\mathbf{0}$. We use concatenation directions (left or right) to indicate different responses, i.e., correct or wrong.

We conduct a similar encoding mechanism for the KS module. Since the KS module only focuses on the general knowledge state changes regardless of question specific variations, the interaction embedding $\mathbf{c}_t$ of KS is 

\vspace{-0.3cm}

\begin{align}
	\label{eq:ct}
	\mathbf{c}_{t}=\left\{
		\begin{array}{l}
			\bar{\mathbf{k}}_t \oplus \mathbf{0}, \quad r_{t}=1 \\
			\mathbf{0} \oplus \bar{\mathbf{k}}_t, \quad  r_{t}=0 \\
		\end{array}\right.
\end{align}

\subsection{Question-centric Knowledge Acquisition Module}
\label{sec:ka_module}
Students absorb knowledge as they interact with questions and their knowledge acquisition varies after solving the homogeneous questions. Hence, we propose to estimate students' question-specific knowledge acquisition with the joint representations $\mathbf{e}_t$s of the questions, concepts and responses. Similar to the standard DKT model, we choose to use the LSTM cell to update the student's question-level knowledge state $\mathbf{a}_t$ after answering each question at timestamp $t$:

\vspace{-0.5cm}

\begin{align*}
\mathbf{i}_{t} & = \sigma\left(\mathbf{W}_1 \cdot \mathbf{e}_t +\mathbf{U}_1 \cdot \mathbf{a}_{t-1}+\mathbf{b}_1\right) \\
\mathbf{f}_{t} & = \sigma\left(\mathbf{W}_2 \cdot \mathbf{e}_t +\mathbf{U}_2 \cdot \mathbf{a}_{t-1}+\mathbf{b}_2\right) \\
\mathbf{o}_{t} & = \sigma\left(\mathbf{W}_3 \cdot \mathbf{e}_t +\mathbf{U}_3 \cdot \mathbf{a}_{t-1}+\mathbf{b}_3\right) \\
\mathbf{\tilde{c}}_{t} & = \sigma\left(\mathbf{W}_4 \cdot \mathbf{e}_t +\mathbf{U}_4 \cdot \mathbf{a}_{t-1}+\mathbf{b}_4\right) \\
\mathbf{c}_{t} & = \mathbf{f}_{t} \odot \mathbf{c}_{t-1}+\mathbf{i}_{t} \odot \mathbf{\tilde{c}}_{t} \\
\mathbf{a}_t & = \mathbf{o}_{t} \odot \tanh \left(\mathbf{c}_{t}\right) 
\end{align*}

\noindent where $\mathbf{W}_i$s, $\mathbf{U}_i$s, $\mathbf{b}_i$s are trainable parameters and $\mathbf{W}_i \in \mathbb{R}^{d \times 4d}$, $\mathbf{U}_i \in \mathbb{R}^{d \times d}$, $\mathbf{b}_i \in \mathbb{R}^{d \times 1}$ and $i = 1, 2, 3, 4$. $\sigma$, $\odot$, and $\tanh$ denote the sigmoid, element-wise multiplication and hyperbolic tangent functions.

Different from existing approaches that directly use (with linear transformation) the learned knowledge state ($\mathbf{a}_t$) to predict the student knowledge mastery, we apply a knowledge acquisition network to first extract the knowledge states with a fully connected neural layer and then project it into the question-centric space via non-linear transformation. The question-centric knowledge acquisition score $\alpha_{t}$ is computed as follows:


\vspace{-0.3cm}
\begin{equation*}
\alpha_{t} = \mbox{S-Pool} \Bigl(\mathbf{w}^a \odot \mbox{ReLU} \bigl( \mathbf{W}^a_2 \cdot \mbox{ReLU} ( \mathbf{W}^a_1 \cdot \mathbf{a}_t + \mathbf{b}^a_1 ) + \mathbf{b}^a_2 \bigl) \Bigl) 
\end{equation*}


\noindent where $\mathbf{W}^a_1$, $\mathbf{W}^a_2$, $\mathbf{w}^a$, $\mathbf{b}^a_1$ and $\mathbf{b}^a_2$ are trainable parameters and $\mathbf{W}^a_1 \in \mathbb{R}^{d \times d}$, $\mathbf{W}^a_2 \in \mathbb{R}^{n \times d}$, $\mathbf{w}^a \in \mathbb{R}^{n \times 1}$, $\mathbf{b}^a_1 \in \mathbb{R}^{d \times 1}$, $\mathbf{b}^a_2 \in \mathbb{R}^{n \times 1}$, $n$ is the total number of questions, and S-Pool means sum pooling.

\subsection{Question-agnostic Knowledge State Module}
\label{sec:ks_module}
In real educational contexts, students may frequently guess or slip when they interact with questions. This may cause the KA module overly sensitive to each interaction and hence lead prediction confusion during the inference stage. Therefore, as an important complement, we model the general question-agnostic changes of students' knowledge state in the KS module. Similar to  knowledge state modeling in the KA module, we apply another LSTM cell to update the student's question-agnostic knowledge state ($\mathbf{g}_t$) after receiving each response. The LSTM cell in the KS module takes question-agnostic input $\mathbf{c}_t$ instead of $\mathbf{e}_t$. The iterative update equations of $\mathbf{g}_t$ are described in Appendix A.1 due to the space limit. Furthermore, we design a knowledge state extraction network to capsule the general knowledge states of a student by applying non-linear transformations to project the mastery level into the space of KCs and computing the knowledge mastery score $\beta_{t}$ as follows:

\vspace{-0.3cm}
\begin{equation*}
\beta_{t} = \mbox{S-Pool} \Bigl( \mathbf{w}^g \odot \mbox{ReLU} \bigl( \mathbf{W}^g_2 \cdot \mbox{ReLU} ( \mathbf{W}^g_1 \cdot \mathbf{g}_t + \mathbf{b}^g_1 ) + \mathbf{b}^g_2 \bigl) \Bigl)
\end{equation*}


\noindent where $\mathbf{W}^g_1$, $\mathbf{W}^g_2$, $\mathbf{w}^g$, $\mathbf{b}^g_1$ and $\mathbf{b}^g_2$ are trainable parameters and $\mathbf{W}^g_1 \in \mathbb{R}^{d \times d}$, $\mathbf{W}^g_2 \in \mathbb{R}^{m \times d}$, $\mathbf{w}^g \in \mathbb{R}^{m \times 1}$, $\mathbf{b}^g_1 \in \mathbb{R}^{d \times 1}$, $\mathbf{b}^g_2 \in \mathbb{R}^{m \times 1}$.

\subsection{Question-centric Problem Solving Module}
\label{sec:ps_module}
Correctly answering a question not only depends on the students' knowledge mastery, but is highly relevant to the question itself such as its difficulty and discrimination. Therefore, we present a question-centric problem solving module to estimate students' knowledge application abilities to specific questions by projecting their knowledge mastery on questions. Specifically, we design a problem solving network to conduct such knowledge projection as follows:

\vspace{-0.5cm}
\begin{align*}
& \mathbf{p}_{t+1} = \mathbf{g}_t \oplus \mathbf{q}_{t+1} \oplus \bar{\mathbf{k}}_{t+1} \\
& \zeta_{t+1} = \mathbf{w}^p \cdot \mbox{ReLU} \bigl( \mathbf{W}^p_2 \cdot \mbox{ReLU} ( \mathbf{W}^p_1 \cdot \mathbf{p}_{t+1} + \mathbf{b}^p_1 ) + \mathbf{b}^p_2 \bigl) + b^p
\end{align*}

\noindent where $\zeta_{t+1}$ denotes the students' knowledge application score on question $q_{t+1}$. $\mathbf{p}_{t+1}$ contains both the student knowledge mastery at time $t$ and all the available information about question $q_{t+1}$ and $\mathbf{p}_{t+1} \in \mathbb{R}^{3d \times 1}$. $\mathbf{W}^p_i$s, $\mathbf{w}^p$, $\mathbf{b}^p_i$s and $b^p$ are trainable parameters and $\mathbf{W}^p_i \in \mathbb{R}^{3d \times 3d}$, $\mathbf{b}^p_i \in \mathbb{R}^{3d \times 1}$,$\mathbf{w}^p \in \mathbb{R}^{1 \times 3d}$, $b^p$ is scalar and $i = 1, 2$.

\subsection{Interpretable Prediction Layer}
\label{sec:prediction}
In general, it is very challenging to explain the DLKT models' parameters and their prediction decision making processes. Therefore, in this work, we design an IRT based prediction layer to enhance the prediction interpretability of the proposed QIKT model. Similar to previous work by \citet{yeung2019deep}, we choose to use the IRT function to calculate the probability that a student will answer the question correctly. Furthermore, we strict the IRT function only takes the linear combined scores of question-centric knowledge acquisition score from the KA module, the knowledge mastery score from the KS module and the knowledge application score from the PS module, defined as follows:


\vspace{-0.2cm}
$$\hat{r}_{t+1} = \sigma(\alpha_t + \beta_t + \zeta_{t+1})$$

Please note that we explicitly choose not to include any learnable parameters inside the IRT based prediction function for better interpretability.

\subsection{Optimization of QIKT}
\label{sec:train}

 All learnable parameters of our QIKT model are optimized by minimizing the binary cross entropy loss between the ground-truth responses $r_i$s and the estimated probabilities $\hat{r}_i$s from the IRT layer as the objective function:

\vspace{-0.2cm}
\begin{equation*}
\mathcal{L}_{\mbox{\tiny IRT}} = - \sum_{i} \bigl( r_i \log \hat{r}_i + (1-r_i) \log (1-\hat{r}_i) \bigl)
\end{equation*}
\vspace{-0.4cm}

Furthermore, to directly improve the discriminative ability of the internal knowledge related scores from the KA, KS and PS modules, we explicitly cast these scores via sigmoid function and add the optimization terms about question-centric knowledge acquisition scores ($\alpha_t$s), knowledge mastery scores ($\beta_t$s) and knowledge application scores ($\zeta_t$s) into the overall model training process. Therefore, the final optimization function is:

\vspace{-0.3cm}
\begin{equation*}
	\mathcal{L} = \mathcal{L}_{\mbox{\tiny IRT}} + \lambda \bigl(\mathcal{L}_{*}(\boldsymbol{\alpha}) + \mathcal{L}_{*}(\boldsymbol{\beta}) + \mathcal{L}_{*}(\boldsymbol{\zeta}) \bigl)
\end{equation*}

\noindent where $\lambda$ is tuning hyper-parameter. $\boldsymbol{\alpha}$, $\boldsymbol{\beta}$, and $\boldsymbol{\zeta}$ denote the collections of the corresponding scores, i.e., $\boldsymbol{\alpha} = \{\alpha_i\}$, $\boldsymbol{\beta} = \{\beta_i\}$, and $\boldsymbol{\zeta} = \{\zeta_i\}$ and $\mathcal{L}_{*}(\mathbf{z})$ is defined as follows:

\vspace{-0.3cm}
\begin{equation*}
\mathcal{L}_{*}(\mathbf{z})  = - \sum_i \bigl( r_i \log \sigma(z_i) + (1-r_i) \log (1-\sigma(z_i)) \bigl)  
\end{equation*}

\section{Experiment}

In this section, we present details of our experiment settings and the corresponding results. We conduct comprehensive analyses and investigations to illustrate the effectiveness of our QIKT model. We have provided the data and code of QIKT along with this submission.

\subsection{Datasets}
\label{sec:dataset}
\begin{table*}[!hptb]
\small
\centering
\begin{tabular}{lcccccc}
\toprule
Dataset & \# of Ss & \# of Is & \# of Qs & \# of KCs & Avg. KCs  & Avg. Qs \\ \hline
\textbf{ASSIST2009}  & 3,852    & 282,605      & 17,737    & 123 & 1.197   & 144.2         \\
\textbf{Algebra2005} & 574     & 607,013      & 173,109   & 112 & 1.364   & 1545.6        \\
\textbf{NeurIPS34}   & 4,918    & 1,382,661     & 948      & 57  & 1.015   & 16.6          \\
\bottomrule
\end{tabular}
\caption{Data statistics of three datasets. \# of Ss/Is/Qs denote the number of students, interactions and questions. \emph{Avg. KCs} and \emph{Avg. Qs} denotes the number of KCs per question and the number of questions per KC.} 
\label{tab:sta}
\vspace{-0.2cm}
\end{table*}

We use three widely used publicly available datasets to evaluate the performance of QIKT:

\begin{itemize}[leftmargin=*]
\item \textbf{ASSISTments2009\footnote{https://sites.google.com/site/assistmentsdata/home/2009-2010-assistment-data/skill-builder-data-2009-2010} (ASSIST2009)}: is collected from ASSISTment online tutoring platform in the school year 2012-2013 that students are assigned to answer similar exercises from the skill builder problem sets. 
\item \textbf{Algebra 2005-2006\footnote{https://pslcdatashop.web.cmu.edu/KDDCup/} (Algebra2005)}: is provided by the KDD Cup 2010 EDM Challenge where students need to complete the steps to achieve the mastery of the related KCs.
\item \textbf{NeurIPS2020 Education Challenge\footnote{https://eedi.com/projects/neurips-education-challenge} (NeurIPS34)}: is released in Task 3 and Task 4 in the NeurIPS2020 Education Challenge that consists of students' answer to multiple-choice diagnostic questions in mathematics \cite{wang2020instructions}.
\end{itemize}

To conduct reproducible experiments, we rigorously follow the data pre-processing steps suggested in \cite{liu2022pykt}. We remove student sequences shorter than 3 attempts. Data statistics are summarized in Table \ref{tab:sta}.

\subsection{Baselines}
\label{sec:baselines}
\begin{table*}[!hptb]
\small
\centering
\begin{tabular}{c|c|c|c|c|lll}
\toprule	
\multirow{2}{*}{Method} & \multirow{2}{*}{Model Type} & \multirow{2}{*}{\begin{tabular}[c]{@{}c@{}}Usage of \\ Questions\end{tabular}} & \multirow{2}{*}{\begin{tabular}[c]{@{}c@{}}Usage of \\ KCs\end{tabular}} & \multirow{2}{*}{\begin{tabular}[c]{@{}c@{}}Is \\ interpretable\end{tabular}} & \multicolumn{3}{c}{AUC}                                       \\  \cline{6-8}
                        &                       &                                                                           &                                                                          &                                                                              & ASSIST2009             & Algebra2005            & NeurIPS34                                                      \\
\hline
DKT                     & Sequential            & No                                                                        & Yes                                                                      & No                                                                           & 0.7541±0.0011*                 & 0.8149±0.0011*                  & 0.7689±0.0002*                  \\
DKT+                    & Sequential            & No                                                                        & Yes                                                                      & No                                                                           & 0.7547±0.0017*                 & 0.8156±0.0011*                  & 0.7696±0.0002*                  \\
KQN                     & Sequential            & No                                                                        & Yes                                                                      & No                                                                           & 0.7477±0.0011*                 & 0.8027±0.0015*                  & 0.7684±0.0003*                  \\
qDKT                    & Sequential            & Yes                                                                       & No                                                                       & No                                                                           & 0.7016±0.0049*                 & 0.7485±0.0017*                  & 0.7995±0.0008*                  \\
DKT-IRT                 & Sequential            & No                                                                        & Yes                                                                      & Yes                                                                          & 0.7591±0.0007*                 & 0.8290±0.0004*                  & 0.7695±0.0004*                  \\
IEKT                    & Sequential            & Yes                                                                       & Yes                                                                      & No                                                                           & 0.7861±0.0027*                 & \textbf{0.8416±0.0014$\bullet$} & \textbf{0.8045±0.0002$\bullet$} \\
DeepIRT                 & Memory                & No                                                                        & Yes                                                                      & Yes                                                                          & 0.7465±0.0006*                 & 0.8040±0.0013*                  & 0.7672±0.0006*                  \\
DKVMN                   & Memory                & No                                                                        & Yes                                                                      & No                                                                           & 0.7473±0.0006*                 & 0.8054±0.0011*                  & 0.7673±0.0004*                  \\
ATKT                    & Adversarial           & No                                                                        & Yes                                                                      & No                                                                           & 0.7470±0.0008*                 & 0.7995±0.0023*                  & 0.7665±0.0001*                  \\
GKT                     & Graph                 & No                                                                        & Yes                                                                      & No                                                                           & 0.7424±0.0021*                 & 0.8110±0.0009*                  & 0.7689±0.0024*                  \\
SAKT                    & Attention             & No                                                                        & Yes                                                                      & No                                                                           & 0.7246±0.0017*                 & 0.7880±0.0063*                  & 0.7517±0.0005*                  \\
SAINT                   & Attention             & Yes                                                                       & Yes                                                                      & No                                                                           & 0.6958±0.0023*                 & 0.7775±0.0017*                  & 0.7873±0.0007*                  \\
AKT                     & Attention             & Yes                                                                       & Yes                                                                      & No                                                                           & 0.7853±0.0017*                 & 0.8306±0.0019*                  & 0.8033±0.0003*                  \\
\hline
QIKT                    & Sequential            & Yes                                                                       & Yes                                                                      & Yes                                                                          & \textbf{0.7878±0.0024}         & 0.8408±0.0007                   & 0.8044±0.0005                  \\
\bottomrule
\end{tabular}
\caption{The overall prediction performance of all the baseline models and our QIKT. We highlight the highest results with bold. Marker $*$, $\circ$ and $\bullet$ indicates whether the proposed model is statistically superior/equal/inferior to the compared method (using paired t-test at 0.01 significance level).}
\label{tab:overall}
\end{table*}

We compare our QIKT with the following state-of-the-art DLKT models to evaluate the effectiveness of our approach:

\begin{itemize}[leftmargin=*]
\item \textbf{DKT}: leverages an LSTM layer to encode the student knowledge state to predict the students' response performances \cite{piech2015deep}.
\item \textbf{DKT+}: an improved version of DKT to solve the reconstruction and non-consistent prediction problems \cite{yeung2018addressing}.
\item \textbf{KQN}: uses student knowledge state encoder and skill encoder to predict the student response performance via the dot product \cite{lee2019knowledge}.
\item \textbf{qDKT}: predicts the future performance of student knowledge state at the question level \cite{sonkar2020qdkt}.
\item \textbf{DKT-IRT}: incorporates IRT to improve the interpretability of DKT \cite{converse2021incorporating}.
\item \textbf{IEKT}: models student knowledge state via the student cognition and knowledge acquisition estimation \cite{long2021tracing}.
\item \textbf{DKVMN}: designs a static key matrix to store the relations between the different KCs and a dynamic value matrix to update the students' knowledge state \cite{zhang2017dynamic}.
\item \textbf{DeepIRT}: a combination of the IRT and DKVMN to enhance the interpretability of memory augmented models \cite{yeung2019deep}.
\item \textbf{ATKT}: performs adversarial perturbations into student interaction sequence to improve DLKT model's generalization ability \cite{guo2021enhancing}.
\item \textbf{GKT}: utilizes the graph structure to predict the student response performance \cite{nakagawa2019graph}.
\item \textbf{SAKT}: utilizes a self-attention mechanism to capture relations between exercises and the student responses \cite{pandey2019self}.
\item \textbf{SAINT}: uses the Transformer-based encoder-decoder architecture to capture students' exercise and response sequences \cite{choi2020towards}.
\item \textbf{AKT}: leverages an attention mechanism to characterize the time distance between questions and the past interaction of students \cite{ghosh2020context}.
\end{itemize}

\subsection{Experimental Setup}
\label{sec:setup}

We set the maximum length of model input sequence to 200 and perform 5-fold cross-validation for every combination of models and datasets. We use 80\% of student sequences for training and validation, and use the rest 20\% of student sequences for model evaluation. We adopt ADAM optimizer to train all the models \cite{kingma2015adam}. The number of training epochs is set to 200. We choose to use early stopping strategy that stops optimization when the AUC score is failed to get the improvement on the validation set in the latest 10 epochs. The hyper-parameter $\lambda$, the learning rate and the embedding size $d$ are searching from [0,0.5,1,1.5,2], [1e-3, 1e-4, 1e-5], [64, 256] respectively. All the models are implemented in PyTorch and are trained on a cluster of Linux servers with the NVIDIA RTX A5000 GPU device. Following all existing DLKT research, we use the Area Under the Curve (AUC) as the main evaluation metric. We also choose to use Accuracy as the secondary evaluation metric.

\vspace{-0.2cm}
\subsection{Results}
\label{sec:results}

\begin{figure*}[!hbpt]
\centering
\includegraphics[width=0.85\textwidth]{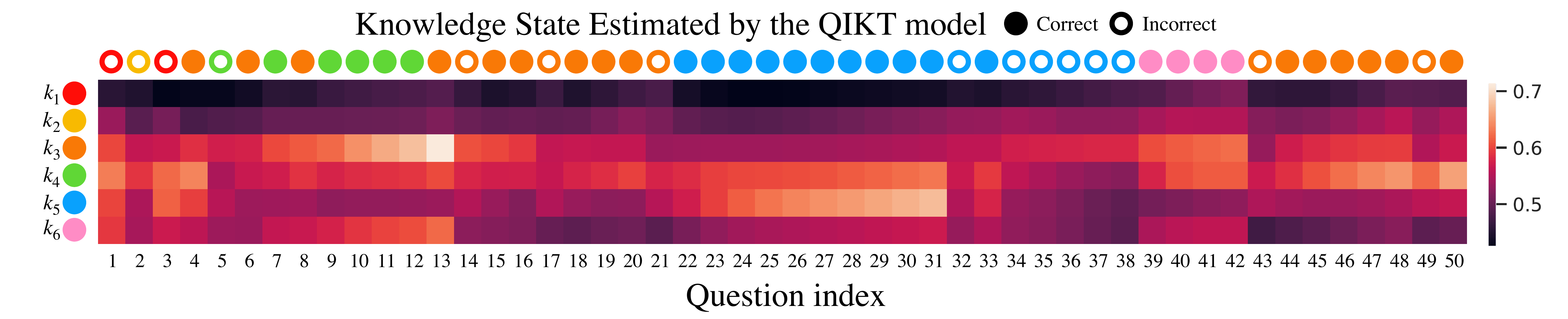}  \vspace{-0.4cm}
\caption{An example of a student knowledge states of 6 concepts as student's solve 50 questions of NeurIPS34.} 
\label{fig:predict_sequence}
\end{figure*}

\noindent \textbf{Overall Performance}. Due to the space limit, results in terms of accuracy and the details of statistical tests are provided in Appendix A.2 and Appendix A.3. The overall model performance is reported in Table \ref{tab:overall}. From Table \ref{tab:overall}, we make the following observations: (1) Our proposed model QIKT significantly outperforms 13 baselines on all three datasets (except we have two loss with IEKT on ASSIST2009 and NeurIPS34 datasets). More importantly, as a representative of the deep sequential KT models, compared with DKT, our proposed model improves the AUC by 3.30\%, 2.60\% and 3.60\% on three datasets. That show our proposed modules can significantly improve the performance. (2) When comparing performance on ASSIST2009, Algebra2005 to NeurIPS34, DLKT models behave quite differently. For example, DKT significantly outperforms qDKT on the ASSIST2009 and Algebra2005 datasets by 5.30\% and 6.60\% but is beaten by qDKT by 3.10\% on the NeurIPS34 dataset. Meanwhile, SAINT performs terrible in ASSIST2009 and Algebra2005 datasets, but is pretty good on NeurIPS34 data. We believe this is because the ASSIST2009 and Algebra2005 datasets are much sparser than NeurIPS34 dataset. As we can see from Table \ref{tab:sta}, the average number of questions per KC is 16.6 in NeurIPS34 dataset, which is much smaller compared to the numbers in ASSIST2009 and Algebra2005 (144.2 and 1545.6) datasets. (3) Results between DeepIRT and DKVMN are very close on three datasets, which empirically shows that  the IRT function won't sacrifice the model prediction ability too much. (4) AKT and IEKT are very strong baselines. Both of them use both question and KC related information, which further empirically verifies the importance of considering question-centric representations when building the DLKT models.

\begin{figure}[!bpht]
\centering
\includegraphics[width=\columnwidth]{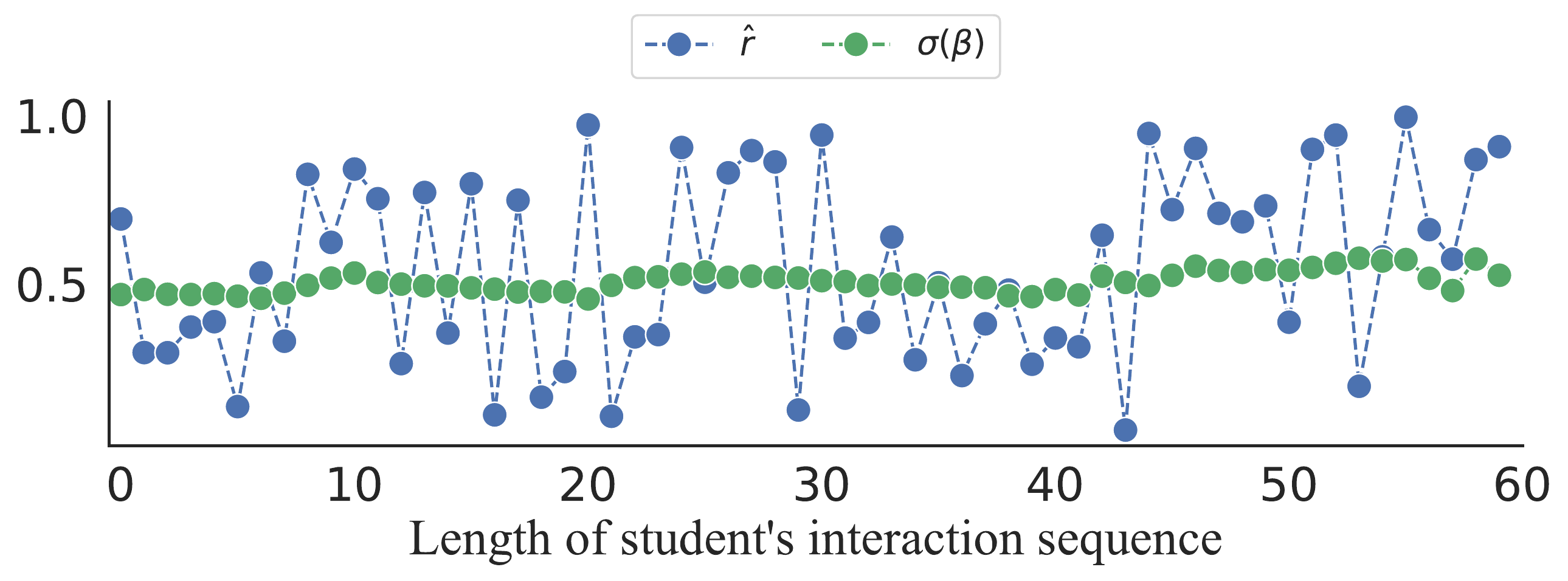}  \vspace{-0.4cm}
\caption{The outputs of QIKT ($\hat{r}$) and module KS ($\sigma{(\beta)}$) at each steps in student's interaction sequence.}
\label{fig:visual_y_1}
\end{figure}

\noindent \textbf{Qualitative Question-centric Effects}. We qualitatively show the question-centric effects of our QIKT model. Figure \ref{fig:visual_y_1} shows the progressive knowledge state estimations of one student with and without question-centric modules, i.e., $\hat{\mathbf{r}}$ v.s. $\sigma(\boldsymbol{\beta})$. As we can see, predictions ($\sigma{(\boldsymbol{\beta})}$) without the question-centric information from the KS module are relatively smooth compared with results from QIKT. We believe this is because the KS module mainly focuses on the knowledge states at KC level, which is insensitive with question variations. When considering both the question-centric knowledge acquisition information and the  question-centric problem solving information, the model outputs distinct predictive results even for the homogeneous questions. Due to the space limit, more illustrative and fine-grained results of $\hat{\mathbf{r}}$, $\sigma(\boldsymbol{\alpha})$,  $\sigma(\boldsymbol{\beta})$ and $\sigma(\boldsymbol{\zeta})$ are provided in Appendix A.4.

\vspace{-0.2cm}
\vspace{-0.3cm}

\noindent \textbf{Interpretable Student Diagnosis}. To verify the interpretable and accurate estimations of students' knowledge states by the proposed QIKT model, we randomly select a student sequence from NeurIPS34 and observe the knowledge state variations of the student in 50 questions with 6 KCs. From Figure \ref{fig:predict_sequence}, we make the following observations: (1) since the student always gives wrong responses to $k_3$ after answering the question 20 (e.g. question 21,43), the knowledge state of $k_3$ is constantly decline. On the other hand, the student always gives right answers to the questions (question 22-31) which are related to the $k_5$ hence the knowledge acquisition of $k_5$ is constantly increasing.  (2) The knowledge states of little-attempted KCs are slightly lower than those of the diligent-attempted KCs. For example, the knowledge state of $k_6$ is relatively lower than others until the student attempts question 39.

\noindent \textbf{Ablation Study}. We systematically examine the effect of key components by constructing four model variants in Table \ref{tab:ab_study}. ``w/o'' means excludes such module from QIKT. From Table \ref{tab:ab_study}, we can easily observe that (1) comparing QIKT and QIKT w/o IRT, we can see that our IRT based interpretable prediction layer is able to make a good enough trade-off between prediction performance and results interpretability. The AUC score of QIKT decreases 0.09\% on the ASSIST2009 dataset and increases 0.81\% and 0.07\% on Algebra2005 and NeurIPS34 datasets. (2) compared to other variants (e.g., QIKT w/o KS, QIKT w/o PS, and QIKT w/o KS \& PS) that have the IRT prediction layer, QIKT obtains the highest AUC score in all cases except QIKT w/o KS in the NeurIPS34 dataset. This suggests that prediction performance degrades when ignoring any type of question-centric information. Thus, it is important to incorporate question information in DLKT models.

\begin{table}[!hptb]
\small
\setlength\tabcolsep{2pt}
\centering
\begin{tabular}{l|lll}
\toprule
         Method       & ASSIST2009             & Algebra2005            & NeurIPS34              \\
        \hline
         QIKT         & 0.7878±0.0024                   & \textbf{0.8408±0.0007}          & 0.8044±0.0005                   \\ \hline
		 w/o IRT      & \textbf{0.7887±0.0017$\bullet$} & 0.8327±0.0005*                  & 0.8037±0.0004*                  \\
         w/o KS       & 0.7813±0.0019*                  & 0.8365±0.0008*                  & \textbf{0.8048±0.0002$\bullet$} \\
         w/o PS       & 0.7822±0.0022*                  & 0.8345±0.0005*                  & 0.8037±0.0002*                  \\
         w/o KS \& PS & 0.7442±0.0043*                  & 0.7487±0.0008*                  & 0.8032±0.0002*                  \\
        \bottomrule
\end{tabular}
\caption{The performance of different variants in QIKT. Marker $*$, $\circ$ and $\bullet$ indicates whether our proposed model is statistically superior/equal/inferior to the compared method (using paired t-test at 0.01 significance level).}
\label{tab:ab_study}
\vspace{-0.5cm}
        
\end{table}

\vspace{-0.3cm}
\section{Conclusions}
\label{sec:conclusion}
In this paper, we propose an interpretable deep sequential KT model learning framework with question-centric cognitive representations. Comparing with existing DLKT models, our QIKT model is able to estimate students' knowledge acquisition and measure the student problem solving ability for each specific question. Furthermore, we design an IRT based interpretable layer to make the QIKT's prediction results more explainable. Quantitative and qualitative experiment results on three real-world datasets demonstrate that QIKT outperforms other state-of-the-art DLKT learning approaches in terms of AUC and is able to generate explainable predictions for tutors and students.

\section*{Acknowledgments}
This work was supported in part by National Key R\&D Program of China, under Grant No. 2020AAA0104500; in part by Beijing Nova Program (Z201100006820068) from Beijing Municipal Science \& Technology Commission; in part by NFSC under Grant No. 61877029 and in part by Key Laboratory of Smart Education of Guangdong Higher Education Institutes, Jinan University (2022LSYS003).

\bibliography{aaai2023.bib}

\newpage
\onecolumn

\end{document}